\documentclass[twocolumn]{article}

\usepackage[utf8]{inputenc}
\usepackage[T1]{fontenc}
\usepackage[a4paper, total={7in, 10in}]{geometry}
\usepackage{graphicx}
\usepackage{xcolor}
\usepackage[authoryear]{natbib}
\usepackage{microtype}
\usepackage[parfill]{parskip}
\usepackage[colorlinks = true,
            linkcolor = purple,
            urlcolor  = blue,
            citecolor = cyan,
            anchorcolor = black]{hyperref}
\usepackage{amsmath,amsthm,amssymb,amsfonts}
\usepackage{booktabs}
\usepackage{nicefrac}
\usepackage{bm}
\usepackage{xpatch}
\usepackage{tikz}
\usepackage[inline,shortlabels]{enumitem}
\usepackage{algorithm}
\usepackage[noend]{algpseudocode}
\usepackage[capitalize,noabbrev]{cleveref}
\usepackage{subcaption}

\usepackage[american]{babel}

\usepackage{natbib} 
\usepackage{mathtools} 
\usepackage{booktabs} 
\usepackage{tikz} 
\usepackage{bbm}
\usepackage{amsfonts}

\graphicspath{{figs/}}

\makeatletter
\xpatchcmd{\algorithmic}{\itemsep\z@}{\itemsep=0.5ex plus1pt}{}{}
\makeatother

\usetikzlibrary{cd,arrows,calc,shapes,positioning}
\tikzstyle{obs} = [circle,fill=white,draw=black,inner sep=0pt,minimum size=18pt,font=\fontsize{10}{10}\selectfont,node distance=1,thick]
\tikzstyle{latent} = [obs,dotted]

\definecolor{lightgrey}{rgb}{0.925, 0.925, 0.925}

\title{The Causal Marginal Polytope for Bounding Treatment Effects}
\date{}

\usepackage{authblk}

\author[1]{Jakob Zeitler}
\author[1]{Ricardo Silva}
\affil[1]{University College London}
\affil[ ]{\small \textit{Corresponding author: jakob.zeitler.19@ucl.ac.uk}}

\begin{document}

\twocolumn[
  \begin{@twocolumnfalse}
  
\maketitle

\begin{abstract}
\noindent Due to unmeasured confounding, it is often not possible to identify causal effects from a postulated model. Nevertheless, we can ask for \emph{partial identification}, which usually boils down to finding upper and lower bounds of a causal quantity of interest derived from all solutions compatible with the encoded structural assumptions. One appealing way to derive such bounds is by casting it in terms of a constrained optimization method that searches over all causal models compatible with evidence, as introduced in the classic work of Balke and Pearl (1994) for discrete data. Although by construction this guarantees tight bounds, it poses a formidable computational challenge. To cope with this issue, alternatives include algorithms that are not guaranteed to be tight, or by introducing restrictions on the class of models. In this paper, we introduce a novel alternative: inspired by ideas coming from belief propagation, we enforce compatibility between \emph{marginals} of a causal model and data, without constructing a \emph{global} causal model. We call this collection of locally consistent marginals the \emph{causal marginal polytope}. As global independence constraints disappear when considering small dimensional tractable marginals, this also leads to a rethinking of how to elicit and express causal knowledge. We provide an explicit algorithm and implementation of this idea, and assess its practicality with numerical experiments.
\end{abstract}

\vspace{1cm}
  \end{@twocolumnfalse}
]

\section{Introduction}
\label{sec:intro}

In causal inference, we are concerned about identifying and estimating properties of a distribution under an \emph{intervention}. Using Pearl's notation \citep{pearl:09}, a quantity such as $\mathbb E[Y~|~do(X = x)]$ describes the expected value of a random variable $Y$ when $X$ is \emph{controlled} to take the value $x$. It differs from the regression relationship $\mathbb E[Y~|~X = x]$, in which $X$ is a random variable and $X = x$ is a particular realization of this random quantity. Those two function will be equal only under some causal structures \citep{pearl:09,sgs:00}, which in general must come from background knowledge or causal discovery assumptions that are not testable without experimental data. When hidden common causes between treatment $X$ and outcome $Y$ cannot be blocked by observed variables, there might be more than one, and perhaps uncountably many, values for $\mathbb E[Y~|~do(X = x)]$ which are compatible with the observed data.

The problem of \emph{partial identification} amounts to providing the set of causal quantities compatible with observable evidence. It will be a non-trivial set under some background knowledge that rules or constraint the relationship among variables. For instance, in the \emph{instrumental variable} setup we have access to an auxiliary variable $Z$ that is associated with $X$ and causally disconnected from $Y$ under an intervention on $X$. Under such a structure, \cite{balke:94} provided an early example on how to provide such a set by finding an upper bound and a lower bound on quantities such as the average treatment effect (ATE), $\mathbb E[Y~|~do(X = 1)] - \mathbb E[Y~|~do(X = 0)]$ for a binary $X$.

The approach gets tight bounds by parameterizing the causal model without ruling out any possible structural equation that can causally link discrete variables, and by solving the problem as a linear program. The number of decision variables, however, quickly becomes unmanageable both in the number of variables and number of categories each (discrete) variable can take, motivating approaches which attempt to derive a subset of constraints \citep[e.g.][]{robin:12,finkelstein:20}.

In this paper, we contribute with a different direction that complements the existing methods for partial identification in discrete distributions. In a nutshell, it relaxes the problem of bounding causal effects by removing constraints. The relaxation, however, follows a pattern that taps in the familiar formulation of \cite{balke:94} without having to invent new sophisticated approaches for deriving such constraints. Instead, it takes a leaf from generalized belief propagation \cite{yedidia:05}, concepts which are already well-established in the uncertainty in artificial intelligence literature. In particular, we show how to cast the problem in terms of small-dimensional marginals of a causal model, enforcing coherence on overlapping marginals. This forms what we call \emph{the causal marginal polytope}. This requires also thinking about the implications on how to express causal knowledge, as conditional independence constraints disappear as we marginalize a graphical model. 

In Section \ref{sec:background} we describe the problem background, and provide further related work. In Section \ref{sec:method}, we describe the main construction of the causal marginal polytope for discrete data, assuming first knowledge of true distributions under particular observational and interventional regimes, and then discussing a basic overview on how to use the method with finite samples. Section \ref{sec:experiments} illustrates the method with numerical examples

\section{Model Setup and Related Work}
\label{sec:background}

We start reviewing the parameterization of discrete distributions according to the setup of \cite{balke:94}. This is contrasted with further extensions and alternative approaches.

\subsection{Parameterizing discrete causal models}

A \emph{structural causal model} (SCM) \citep{pearl:09} describes a system of observed variables $\mathbf V \equiv \{V_1, V_2, \dots, V_p\}$ as causally connected by a set of \emph{structural equations} $\{f_1, f_2, \dots, f_p\}$ and \emph{background} variables $\mathbf U \equiv \{U_1, U_2, \dots, U_p\}$ following a distribution. That is, for each $V_i \in \mathbf V$, we have a function $f_i$ such that
\[
V_i = f_i(V_{pa_i}, U_i),
\]
\noindent where $V_{pa_i} \subset \mathbf V$ are called the (observable) parents of $V_i$, in the sense that we can construct a directed acyclic graph (DAG) among the elements of $\mathbf V$ with the corresponding set of parents following the usual definitions from graph theory.

We assume there is a joint distribution $F_\mathbf U$ of $\mathbf U$, keeping in mind that in principle each $U_i$ could be infinite-dimensional, potentially forming a collection of stochastic processes. However, when all variables in $\mathbf V$ are discrete, it is possible to assume without loss of generality that each $U_i$ can be mapped to a particular value of a scalar categorical variable $R_i$ with finitely many values, as originally introduced by \cite{balke:94}. In particular, considering the case where all variables in $\mathbf V$ are binary, any structural causal model can be represented as
\[
V_i = f_{R_i}(V_{pa_i})),
\]
\noindent where $R_i \in \{1, 2, \dots, 2^{2^{|pa(V_i)|}}\}$ follows a categorical distribution. The finite function space $\{f_{r_i}\}$ spans all functions in $\{0, 1\}^{|pa(V_i)|} \rightarrow \{0, 1\}$. The joint distribution of $R_1, R_2, \dots, R_p$ can incorporate probabilistic constraints such as marginal independence. It is commonly the case that we represent a model of marginal independence using a \emph{bidirected graphical model}, where each pair of vertices is either connected by a bidirected edge or has no edge. The global Markov structure of such models describes that any two sets of vertices $R_{S_a}$ and $R_{S_b}$, for which no bidirected edge exists across them, should have a corresponding marginal independence structure, that is, $p(r_{S_a}, r_{S_b}) = p(r_{S_a})p(r_{S_b})$. When combined with the DAG structure implied by the structural equations, the result is an acyclic directed mixed graph model (ADMG), with Markovian properties described by \cite{richardson:03}. In this case, the ADMG with vertex set $\mathbf V$ includes bidirected edge $V_i \leftrightarrow V_j$ if and only if the bidirected graph model for $\mathbf R$ contains the edge $R_i \leftrightarrow R_j$.

The lack of a bidirected edge is commonly used to encode lack of unmeasured confounding under particular conditioning sets. For instance, if the ADMG implies that some set $V_{B}$ satisfies the backdoor criterion \citep{pearl:09} for a pair $(V_i, V_j)$, then $p(v_j~|~do(v_i), v_B) = p(v_j~|~v_i, v_B)$.

Any interventional distribution can be calculated given $\mathcal F$, the function spaces of structural equations for all observed variables, and the distribution of the background distribution $p(r)$. For instance, if we divide $\mathbf V$ into two sets $\mathbf V_A$ and $\mathbf V_B$, we have:
\begin{equation}
\begin{array}{c}
\displaystyle p(v_A^\star~|~do(v_B^\star)) = \\
\displaystyle
\sum_r p(r)
\prod_{V_i \in \mathbf V_A} \mathbbm{1}(f_{r_i}(v_{pa_i}) = v_i^\star)
\prod_{V_j \in \mathbf V_B} \mathbbm{1}(v_j = v_j^\star),
\end{array}
\label{eq:implied}
\end{equation}
\noindent where $\mathbbm{1}(\cdot)$ is the indicator function. Information propagates from parents to children given $r$ as the starting point, that is, the value of vector $r$ deterministically implies $v$.

With $\mathcal F$ usually fixed, the only free parameter is the joint distribution $p(r)$. Provided a collection $\mathcal P$ of $k$ distributions in $\mathbf V$, 
\[
\mathcal P \equiv \{p(v_{\backslash D_1}~|~do(v_{D_1})), \dots, p(v_{\backslash D_k}~|~do(v_{D_k}))\},
\]
\noindent where $\mathbf V_{D_k} \subset \mathbf V$ and $\mathbf V_{\backslash D_k} \equiv \mathbf V \backslash \mathbf V_{D_k}$, we say that $p(r)$ is \emph{partially identifiable} from $\mathcal P$ if it is a strict subset of all distributions $\mathcal R$ that are allowed a priori, such as the space of all distributions which are Markov with respect to some bidirected graph.

\subsection{Related work}

Restrictions in $\mathcal F$ and in $\mathcal R$ lead to restrictions in $\mathcal P$. This can lead to model testability, meaning falsification of the restrictions in $\mathcal F$ and $\mathcal R$. Early work includes falsifying discrete instrumental variable models \citep{pearl:95b}. Most restrictions consist of independence constraints corresponding to lack of directed or bidirected edges in the corresponding ADMG of the model, although particular functional constraints such as monotonicity are also sometimes used.

Many methods exist for deriving the constraints in $\mathcal P$ that are implied by $\mathcal F$ and $\mathcal R$. Some common techniques include deriving symbolically the polytope describing distributions in $\mathcal P$ using algebraic methods \citep{balke:95, dawid:03}, but the computational cost of such methods is prohibitive for all but the smallest models. Alternatives include non-exhaustive methods that infer a subset of constraints by a clever combination of graph theory and algebra \citep[e.g.][]{robin:12,finkelstein:20}, or exhaustive and still expensive approaches that exploit independence constraints (lack of edges) for classes of constraints such as equalities \citep{tian:02,evans:18} and inequalities \citep{navascues:20}. Deriving such constraints allows bounding of causal quantities, such as the average treatment effect (ATE).

Numerical approaches express the problem directly by optimizing a quantity of interest such as the ATE (or even a constant, if all we want is to test feasibility) subject to equating the right-hand side of all instances of Eq. \eqref{eq:implied} to (an estimate of) its left-hand side. When marginal independence constraints are to be imposed in $p(r)$, parameterizations such as \cite{drton:08}, or alternative ones such as \cite{duarte:21}, require nonlinearities. As discussed by \cite{duarte:21}, however, such parameterizations are polynomial programs still amenable to some guarantees of optimality, and allow for the use of efficient implementations of mathematical programming packages.

It still is the case that, even in sparse graphs, the number of decision variables in the mathematical program are in general far too large. One alternative is to constrain the function space $\mathcal F$. In particular, one can postulate that we have a latent variable model with, say, one latent variable per clique in the corresponding bidirected graph of background variables $\mathbf U$, along with non-deterministic conditional probabilities $p(v_i~|~v_{pa_i}, u)$. The model is unidentifiable, but a Bayesian posterior is still well-defined on its parameter space \citep{chickering:96}. The posterior, however, is on the unidentifiable causal parameters and hence cannot be interpreted as providing uncertainty on the bounds \citep{silva_evans:16}. Although by construction the posterior will always be within the bounds that an optimization method can obtain (in the limit of infinite data), barring algorithmic challenges such as the bad mixing of a Markov chain Monte Carlo approach, it will in general be overconstrained, sometimes much narrower than the result of a bounding procedure. How overconstrained will depend on the prior distribution and, even if it correctly captures the estimand of interest, it will be unwarranted if the prior does not follow from acceptable background knowledge about the domain. The shape of the e.g. ATE posterior within the bounds is completely independent of the data (the likelihood being flat in that region), and will not be meaningful unless the prior is informative \citep{silva_evans:16}.

Bounding methods for continuous variables will in general require constraints in (the possibly infinite-dimensional) $\mathcal F$. See \cite{gunsilius2018testability, gunsilius2020, kilbertus:2020,hu:2021,xia:2021} for examples. All of these methods rely heavily on Monte Carlo-powered optimization, special algorithmic tricks to stabilize them, and particular choices of function space. As we are specializing our approach to discrete data, they are out of scope.

\section{The Causal Marginal Polytope}
\label{sec:method}

Our approach is meant to complement ways of parameterizing $\mathcal F$ and $\mathcal R$ in a way to scale up mathematical programs for bounding causal effects. In particular, we will focus on the case where \emph{there is no sparsity}: we will not consider graphs with missing edges to avoid a discussion of ways of parameterizing $p(r)$ \citep[e.g., two alternatives are discussed by][]{drton:08,duarte:21}. Instead, we will provide a way of getting a fully linear program by constraining the ``strength'' of particular edges as motivated by \cite{ramsahai:12} and \cite{silva_evans:16}. We will assume all variables are binary to simplify presentation. We also assume for now that we know the probability distributions $\mathcal P$ from an oracle, while in Section \ref{sec:learning} we discuss what to do to estimate them from data.

\subsection{Model Space and Data Binding}

Let $\mathcal M$ be a collection of (conditional) \emph{marginal causal models} over subsets of set $\mathbf V$, variables which are assumed to follow a SCM $\mathcal M_{global}$. We define a SCM $\mathcal M_j \in \mathcal M$ over a subset $\mathbf V_{M_j} \subset \mathbf V$. Moreover, we also formulate the framework by allowing conditioning on realizations of another subset $\mathbf V_{M'_j} \subset \mathbf V$, which contains only non-descendants of $\mathbf V_{M_j}$ in the causal graph of $\mathcal M_{global}$ (to be clear, in our numerical experimentation in Section \ref{sec:experiments}, we will not make use of conditioning sets $\mathbf V_{M_j}$ in order to focus on other comparative aspects of the model space). 

Model $\mathcal M_j$ is assumed to have a function space $\mathcal F_{\mathcal M_j}$, which follows the same ancestral relationships as those implied by $\mathcal M_{global}$, and a background probability mass function $p_{\mathcal M_j}(r_{M_j}~|~v_{M'_j})$. The background pmf is parameterized by $\theta_{j.v_{M'_j}}$, as a contingency table for each instantiation $v_{M'_j}$ that we include in $\mathcal M_j$. Not all possible instances of the sample space of $\mathbf V_{M'_j}$ need to be defined by $\mathcal M_j$: a reduced number of values $v_{M'_j}$ can be selected according to computational bottlenecks at the expense of introducing further relaxations to the mathematical program. We call the set of possible values $v_{M'_j}$ defined by $\mathcal M_j$ its \emph{conditional scope}.

In the examples of Section \ref{sec:experiments}, the function space $\mathcal F_{\mathcal M_j}$ is exhaustive with respect to $\mathcal M_j$, i.e., it contains all possible corresponding binary structural equations, as in \cite{balke:94}, among the causally-ordered elements of $\mathbf V_{M_j}$. Any variable in $\mathbf V_{M'_j}$ in $\mathcal M_j$ will play a role of ``background'' variable similar to the original definition of $\mathbf U$. This means that the (linear) mapping from $(\mathcal F_{\mathcal M_j}, \theta_{j.v_{M'_j}})$ to $p_{\mathcal M}(v_{\backslash D_{M_j}}~|~v_{M'_j}, do(v_{D_{M_j}}))$, $\mathbf V_{D_{M_j}} \subset \mathbf V_{M_j}$, $\mathbf V_{\backslash D_{M_j}} \equiv \mathbf V_{M_j} \backslash \mathbf V_{D_{M_j}}$, can be done in an entirely analogous way to Eq. \eqref{eq:implied}, with $p_{\mathcal M}(r_{M_j}~|~v_{M'_j})$ taking the role of the averaging distribution. 

The collection of marginals $\mathcal M$ forms a \emph{realizable} set of distributions if there exists a collection of joint distributions of which they are marginals. When we enforce that each $p_{\mathcal M}(\cdot~|~\cdot)$ in $\mathcal M$ must match the corresponding $p(\cdot~|~\cdot)$ in $\mathcal P$, we guarantee that they are realizable for the regimes in $\mathcal P$. The fact that this collection is indexed by different regimes forms a generalization of the \emph{marginal polytope} construction \citep{wain:08}, as usually exploited in the variational inference literature. We will call our collection the \emph{ causal marginal polytope}. As in belief propagation and their generalizations \citep{yedidia:05}, we will relax the problem of defining the set of all causal models compatible with observation by enforcing only \emph{local coherence} over pairwise marginal distributions in $\mathcal M$: submarginals jointly implied by pair $(\mathcal M_{j_a}, \mathcal M_{j_b})$ must agree in each possible causal regime. In particular, they must agree on regimes that are not in $\mathcal P$, as those will be directly matched to each $\mathcal M_j$, a step made more explicit in the next section.

\subsection{Marginalization Constraints and Implications to Modeling}

Our linear program formulation contains the constraint that 
\begin{itemize}
    \item for each $\mathcal M_j \in \mathcal M$, we have that $\theta_{j.{v_{M'_j}}}$ must be in the probability simplex;
    \item for each $\mathcal M_j \in \mathcal M$ where $v_{M'_j}$ is in its conditional scope, we have that $$p_{\mathcal M_j}(v_{\backslash D_{M_j}}~|~v_{M'_j}, do(v_{D_{M_j}})) =$$ $$p(v_{\backslash D_{M_j}}~|~v_{M'_j}, do(v_{D_{M_j}})),$$ if $p(v_{\backslash D_{M_j}}~|~v_{M'_j}, do(v_{D_{M_j}}))$ is contained in $\mathcal P$.
\end{itemize}

Now, consider a pair of causal marginals $\mathcal M_{j_a}$ and $\mathcal M_{j_b}$ where $v_c$ is in the conditional scope of both models, and where $\mathbf V_{M_{j_a}}$, $\mathbf V_{M_{j_b}}$ are the corresponding causal variables in each marginal. Let $\mathbf V_{M_{j_{ab}}} \equiv \mathbf V_{M_{j_a}} \cap \mathbf V_{M_{j_b}}$. Likewise, let $\mathbf V_{D_{M_{j_{ab}}}} \subset \mathbf V_{M_{j_{ab}}}$, with $\mathbf V_{\backslash D_{M_{j_{ab}}}} \equiv \mathbf V_{M_{j_{ab}}} \backslash V_{D_{M_{j_{ab}}}}$ defined accordingly. We call \emph{local coherence} constraints those of the type $$p_{\mathcal M_{j_a}}(v_{\backslash D_{M_{j_{ab}}}}~|~v_{M'_j}, do(v_{D_{M_{j_{ab}}}})) =$$ $$p_{\mathcal M_{j_b}}(v_{\backslash D_{M_{j_{ab}}}}~|~v_{M'_j}, do(v_{D_{M_{j_{ab}}}})).$$

These constraints will be non-redundant if $p(v_{\backslash D_{M_{j_{ab}}}}~|~v_{M'_j}, do(v_{D_{M_{j_{ab}}}}))$ is not in $\mathcal P$.

One implication of working with marginal constraints, besides being a relaxation that trades-off tractability for looser bounds, is that some conditional independencies cannot be explicitly represented. For instance, if we postulate that some observable $X_1$ removes confounding between some $X_2$ and some $X_3$, then a model that excludes $X_1$ while including $X_2$ and $X_3$ will need to express unmeasured confounding between the latter two.

The upside, and an opportunity, is that domain experts may feel more comfortable about expressing knowledge on the \emph{strength} of direct causal effects and unmeasured confounding on smaller marginals of a causal system, than on a full causal structure among all variables. The idea of composite likelihood, for instance, is to express only particular marginals of a likelihood function. Although in machine learning that is mostly motivated by computational tractability, one appeal in statistical modeling is that this reduces exposure to model misspecification \citep{varin:11}.

In the context of partial identification methods, \cite{ramsahai:12} and \cite{silva_evans:16} discuss ways by which ``weak directed effects'' and ``weak unmeasured confounding'' can be encoded. Those constraints were introduced in the context of non-counterfactual approaches for instrumental variable models and are hard to generalize for other structures. Here, we show a way of encoding them in the context of Balke and Pearl's parameterization.

\paragraph{``Weak'' direct effects.} Assume that we elicit from an expert or algorithm that the direct effect of some $X_j$ into $X_i$ is ``weak,'' within the system covered by marginal model $\mathcal M$ and conditional value $V_{\mathcal M'} = v_{\mathcal M'}$. This knowledge must be stated as
$$
|P_{\mathcal M}(V_i = 1~|~do(v_{pa_i \backslash j}), do(V_j = 1), v_{\mathcal M'}) -$$
$$P_{\mathcal M}(V_i = 1~|~do(v_{pa_i \backslash j}), do(V_j = 0), v_{\mathcal M'})| \leq \epsilon_{ij},$$
for a particular parent $V_j$ of $V_i$ in marginal model $\mathcal M$, where $\mathbf V_{pa_i \backslash j}$ is the set of parents of $V_i$ other than $V_j$ and $\epsilon_{ij}$ is a parameter to be elicited from an expert. See \cite{silva_evans:16} for a discussion on its choice, including its derivation from generalizations of structure learning methods, which is out of scope of this manuscript. As $P_{\mathcal M}(V_j = 1~|~do(v_{pa_i}))$ is a linear function of the parameters of $\mathcal M$, per Eq. \eqref{eq:implied}, and the above is a linear function of $P_{\mathcal M}(V_j = 1~|~do(v_{pa_i}))$, the addition of such constraints still imply a linear programming formulation if the objective function is also linear (which will be the case for the ATE). 

Such constraints are expressed as interventional, as opposed to counterfactual: they are testable from experiments even if the parameterization could be interpreted as also implying the existence of cross-world potential outcomes, a criticism of \cite{balke:94} raised by \citep{ramsahai:12}. The case $\epsilon_{ij} = 0$ corresponds to a conditional independence constraint in the experimental distribution where we simultaneously intervene on all parents of $V_i$, and a common interpretion of the lack of an directed edge in a causal DAG.

\paragraph{``Weak'' ancestral bidirected chains.} Start by assuming that we elicit from an expert or algorithm that the level of unmeasured confounding between some some ancestor $X_j$ of $X_i$, and $X_i$ itself, is ``weak'' within the system covered by marginal model $\mathcal M$ and conditional value $V_{\mathcal M'} = v_{\mathcal M'}$. We first consider constraints of the type
$$
|P_{\mathcal M}(V_i = 1~|~do(v_{pa_{ij}}), do(V_j = v_j), v_{\mathcal M'}) -$$
$$P_{\mathcal M}(V_i = 1~|~do(v_{pa_{ij}}), V_j = v_j, v_{\mathcal M'})| \leq \epsilon^\mathsf{C}_{ij},$$
for $v_j = 0, 1$, and where $\mathbf V_{pa_{ij}}$ is the union of the parents of $V_i$ and $V_j$ in $\mathcal M$, other than $\{V_i, V_j\}$ themselves. Once again, $\epsilon^\mathsf{C}_{ij}$ is part of the model input specification.

To avoid formulating a non-linear constraint, we could first attempt to express the second term in the above as
\[
\displaystyle
\frac{
P_{\mathcal M}(V_i = 1, V_j = v_j~|~do(v_{pa_{ij}}), v_{\mathcal M'})}{P(V_j = v_j~|~do(v_{pa_j}), v_{\mathcal M'})},
\]
That is, the denominator can be treated as a constant, \emph{if} the regime where we intervene in $V_{pa_j}$ is in the data. 

For instance, assume that the graph of the marginal is $V_1 \rightarrow V_2 \rightarrow V_3$, with ``unconstrained bidirected edges'' $\{V_1 \leftrightarrow V_2, V_1 \leftrightarrow V_3\}$ and a ``weak'' amount of unmeasured confounding left between $V_2$ and $V_3$. Assume that we have in $\mathcal P$ the regimes corresponding to $do(X_1 = 1)$ and $do(X_1 = 0)$, and that the conditional scope is empty. We then express a ``weak'' $V_2 \leftrightarrow V_3$ as 
$$
|P_{\mathcal M}(V_3 = 1~|~do(v_1), do(V_2 = v_2)) -$$
$$P_{\mathcal M}(V_3 = 1~|~do(v_1), V_2 = v_2)| \leq \epsilon^\mathsf{C}_{ij}.$$

The second term can be expressed as
\[
\displaystyle
\frac{P(V_3 = 1, V_2 = v_2~|~do(v_1))}{P(V_2 = v_2~|~do(v_1))},
\]
\noindent which is given by $\mathcal P$, and hence a constant, making the constraint linear in the parameters. However, if regime $do(v_1)$ is \emph{not} in $\mathcal P$, then the denominator is a function $P_{\mathcal M}(V_2 = v_2~|~do(v_1))$ of the parameters, and the constraint becomes \emph{polynomial}, as in \cite{duarte:21}.

However, once we give up on hard unmeasured confounding constraints (i.e., lack of bidirected edges), we may instead enquire knowledge about the strength of \emph{bidirected paths} that confound $V_i$ and $V_j$. Consider the constraint
\[
|P_{\mathcal M}(V_i = 1~|~do(v_{pa_i}), do(V_j = v_j), V_{\prec j} = v_{\prec j}, v_{\mathcal M'}) -
\]
\[
P_{\mathcal M}(V_i = 1~|~do(v_{pa_i}), V_j = v_j, V_{\prec j} = v_{\prec j}, v_{\mathcal M'}) \leq \epsilon^\mathsf{C}_{ij},
\]
\noindent for $v_j \in \{0, 1\}$, where $\mathbf V_{\prec_j}$ are the ancestors $V_j$ in $\mathcal M$.  For $\epsilon^\mathsf{C}_{ij} = 0$, this would \emph{not} necessarily correspond to the lack of bidirected edge $V_j \leftrightarrow V_i$. In the previous example, without the edge $V_1 \rightarrow V_3$, expressing $P(V_3 = 1~|~do(V_2 = v_2)) \approx P(V_3 = 1~|~V_2 = v_2)$ captures the idea of weak confounding. However, if edge $V_1 \rightarrow V_3$ exists, then we are allowing for a possible backdoor path $V_3 \leftarrow V_1 \rightarrow V_2$. An alternative is check whether input knowledge allows for
$$P(V_3 = 1~|~do(V_2 = v_2), V_1 = v_1) \approx$$ 
$$P(V_3 = 1~|~V_2 = v_2, V_1 = v_1).$$
This would correspond to a constraint on the contribution of the active backdoor path $V_3 \leftrightarrow V_1 \leftrightarrow V_2$ given $V_1$, a ``weak bidirected chain'' that passes through the ancestors of $V_j$. The convenience of this constraint is that $P(V_3 = 1~|~V_2 = v_2, V_1 = v_1)$ is in $\mathcal P$, and hence can be treated as a constant, making the above constraint linear in the parameters. 

Its interpretation is that the ancestors of $V_j$ are taken as a ``approximate covariate adjustment set,'' that is, those ancestors would approximately block confounders between $X_i$ and $X_j$. Notice that confounding induced by vertices ``downstream'' of $X_j$ are accounted by the intervention on the parents of $X_i$. That is, if the structure is $X_2 \rightarrow X_3 \rightarrow X_4$, with bidirected edges $X_2 \leftrightarrow X_3$, and $X_2 \leftrightarrow X_4$, then a statement of ``weak confounding'' between $X_2$ and $X_4$ removes the confounding contribution of the path $X_2 \leftrightarrow X_3 \rightarrow X_4$ via the intervention $do(x_3)$. If desired, we may wish to condition on a subset of the ancestors of $V_j$ as opposed to $V_{\prec j}$.

\subsection{Learning}
\label{sec:learning}

In our illustrative examples of the next section, our applications will be simple enough so that each element of $\mathcal P$ can be estimated just using the frequencies in the data. We do not discuss measures of uncertainty. One idea is discussed by \cite{duarte:21}, treating elements of $\mathcal P$ themselves as unknown, and taking values within a confidence interval obtained by a black-box fitting of the marginals. Likewise, the Bayesian approach  \cite{silva_evans:16} can be immediately used here: a sample from the posterior of a black-box model is passed through the optimization problem (which is, after all, just a functional of $\mathcal P$) resulting in a posterior sample of an objective function of interest.

\subsection{Summary}

To summarize, exploiting a causal marginal polytope has major computational advantages when we want to avoid constraining the function space of structural equations while retaining tractability. The downsides are: i) it is a relaxation, which in general will imply looser bounds - but unless we are willing to introduce untestable constraints in the function space, it is a price we need to pay; ii) given that marginals imply a removal of conditional independencies, we need to rethink which structural knowledge should be enquired. To maintain linearity of the mathematical program, we suggest weak directed edges and weak bidirected chains. This is not necessarily a downside: we may never believe sparsity anyway, even in the full causal system, and it may actually be easier to extract knowledge from enquiring experts about small subsystems than a complex causal system of many variables. If such a collection of marginal models is inconsistent, in the sense that no full causal model is compatible with them, we can still detect where local coherence breaks down as the linear program solver should flag infeasibility.

\section{Experiments}
\label{sec:experiments}

Our goal in this section is to provide numerical examples of an implementation of the causal marginal polytope in action. Training data is considered to be large enough so that, as the optimization is exact, whether we trap or not the true causal effect will be a function of the input knowledge. Hence, we illustrate how informative (or not) bounds turn out to be, as we vary assumptions about weaknesses of edges, as well as our choice of marginals.

One of the primary goals of this section is to highlight how knowledge elicitation in this marginal setting may lead to logical but potentially counter-intuitive results. For instance, it is to be expected that bounds obtained by using 3-dimensional marginals of a 4-dimensional causal model would be at least as wide as, and in fact contain, the bounds found by running a linear program on the 4-variable model. However, this is predicated on the 3-dimensional models being submodels of the corresponding marginal in the 4-variable global model. This would be the case if knowledge were to be specified only in terms of the global Markov properties. However, besides the computational gains, much of the modeling motivation for this work is that sparsity constraints are not necessarily desirable, and that knowledge elicitation from margins is an asset to be added to the modeling toolkit of a practitioner. Model spaces are a function of the restriction parameters $\{\epsilon_{ij}, \epsilon_{ij}^{\mathsf C}\}$. 
We should not expect that human expert judgement calls will entail that restriction parameters chosen from a \textit{marginal} perspective will imply a larger feasible region to the linear program compared to restriction parameters in the \textit{global} model elicited from the same individual (if this was even possible to compute).
Being allowed to think in terms of margins is a contribution that we believe will be of practical value for experts.

\subsection{Setup}

\begin{figure}
  \centering
  \includegraphics[width=1\linewidth]{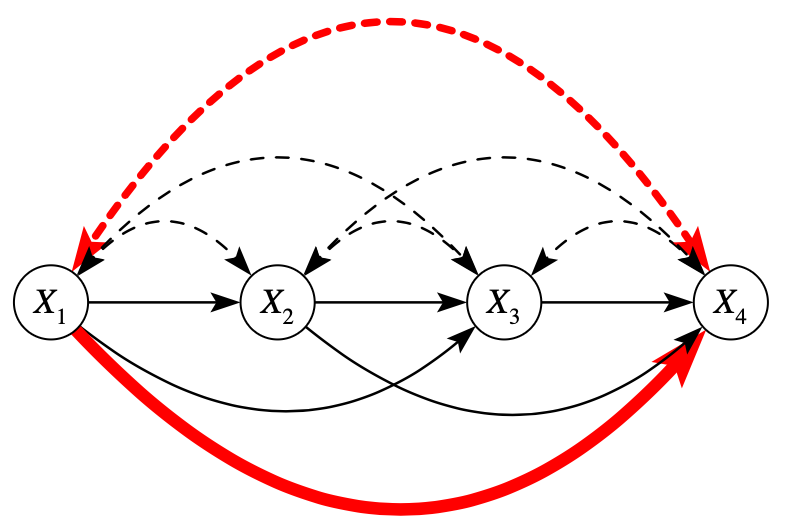}
  \caption{$\mathcal M_{global}$: The red bold directed and bidirected edges  will be constrained by expert knowledge about the causal system.}\label{fig:n4_DAG}
\end{figure}

\begin{figure}
  \centering
  \includegraphics[width=1\linewidth]{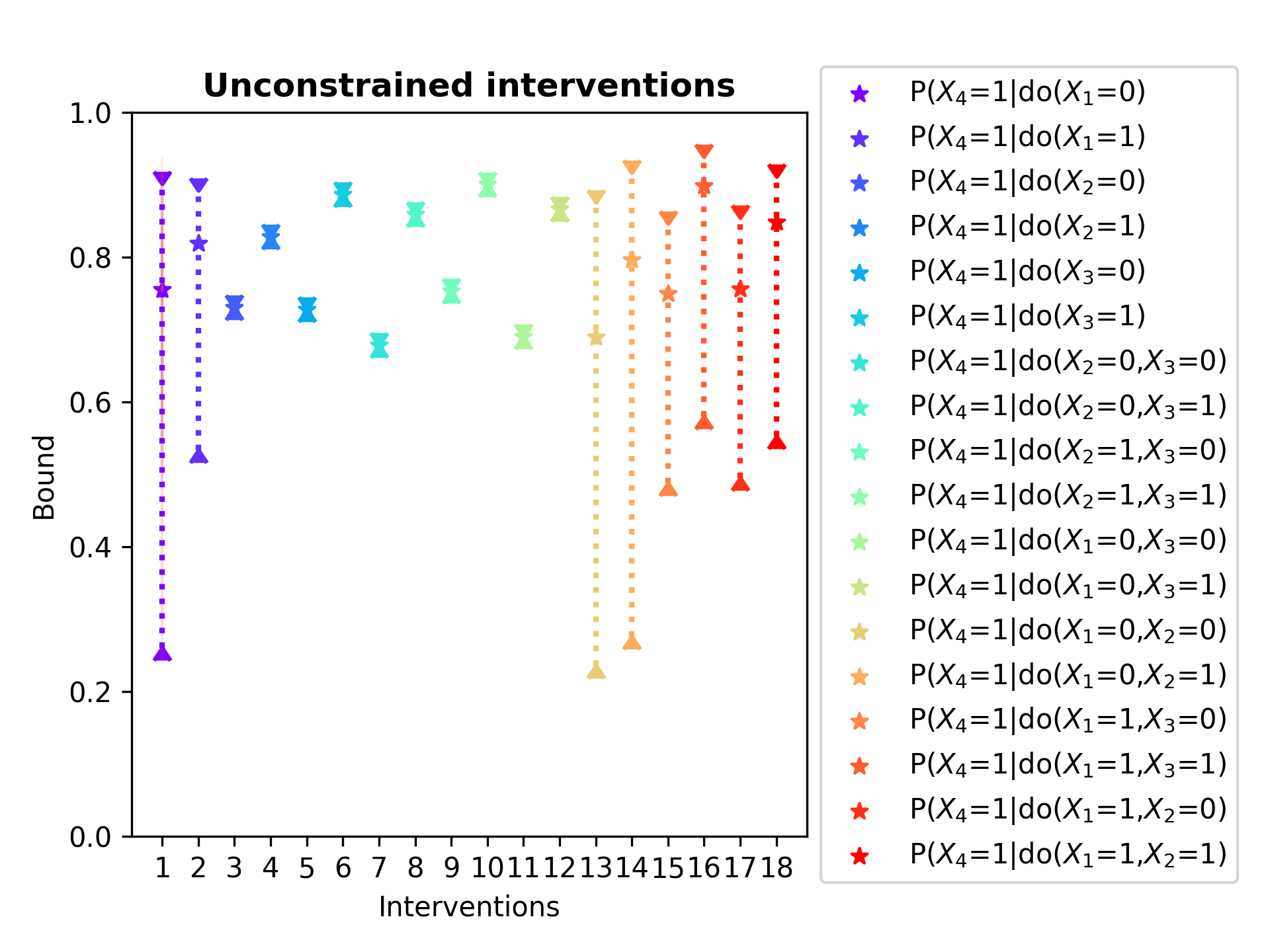}
  \caption{Unconstrained bounds on all single and double interventions. True effects are marked with a star. Intervention 3 to 12 are identified from just data alone.}\label{fig:Unconstrained}
\end{figure}

\textbf{Data} As our starting point for $\mathcal M_{global} = \{X_1, X_2, X_3, X_4\}$ we assume the DAG in Figure \ref{fig:n4_DAG}, i.e. a fully connected graph with full confounding. We construct a simulation from which we can sample a single example SCM with random structural equations. We provide the observational data regime as well as a subset of all possible single and double intervention data regimes to our bounding model: $do(\emptyset),do(x_2),do(x_3),do(x_2,x_3),do(x_1=0,x_3=0),do(x_1=0,x_3=1)$ As such, some interventions will be identifiable, and some will be partially informed from overlapping interventional information or the observational regime, see Figure \ref{fig:Unconstrained}.

\textbf{Model} We choose margins of length 3 from  $\mathcal M_{global}$, i.e. $M=\{$
$M_1=\{X_1,X_2,X_3\},$
$M_2=\{X_1,X_2,X_4\},$
$M_3=\{X_1,X_3,X_4\},$
$M_4=\{X_2,X_3,X_4\}\}$
We enforce the overlap constraint on the overlap sets $ \{ \{X_1,X_2\}, \{X_2,X_4\}, \{X_1,X_3\}, \{X_3,X_4\}\}$. Given assumptions from Figure \ref{fig:n4_DAG}, we then enforce a weak edge $X_1 \rightarrow X_4$ and edge $X_1 \leftrightarrow X_4$ in margins $M_2$ and $M_3$, as $M_1$ and $M_4$ do not contain both $X_1$ and $X_4$.

\subsection{Results}

We will first consider the impact of the constraints individually, then together. True intervention effects are shown as dotted lines. Interventions that are not affected by the constraints are not shown. Figure \ref{fig:Unconstrained} shows the bounds and true intervention effects for all possible single and double interventions of the unconstrained causal marginal polytope.

\begin{figure}
  \centering
  \includegraphics[width=1\linewidth]{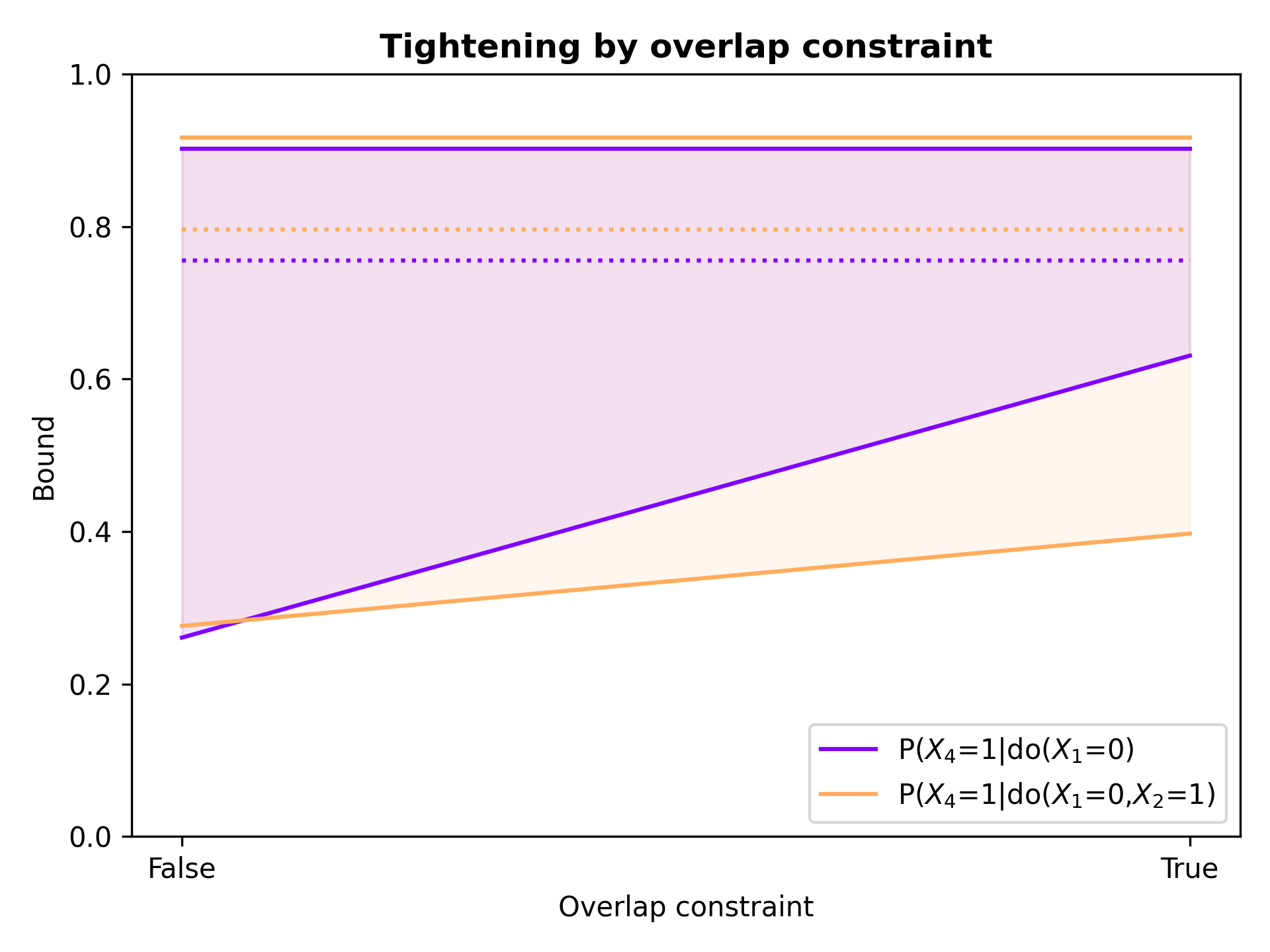}
  \caption{Tightening of bounds by constraining the overlap of the margins.}\label{fig:Overlap}
\end{figure}

\textbf{Overlap} Figure \ref{fig:Overlap} shows the impact of the overlap constraints on bounds, i.e. in this case tightening the lower bound on $P(X_4=1|do(X_1=0,X_2=1)$ from 0.3 to 0.4, and $P(X_4=1|do(X_1=0)$ from 0.3 to 0.6.

\begin{figure}
  \centering
  \includegraphics[width=1\linewidth]{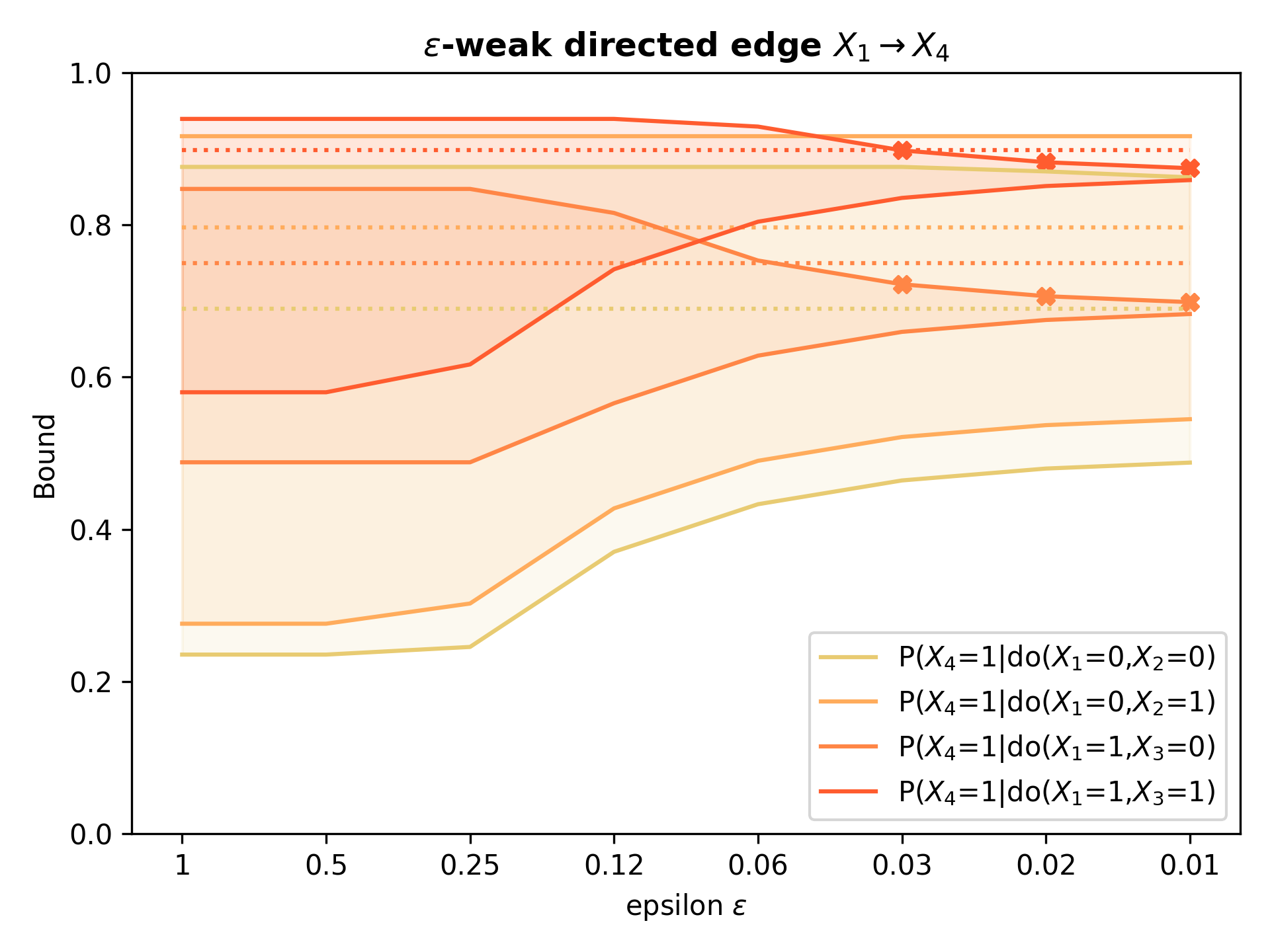}
  \caption{A weak directed edge tightens bounds. Invalid bounds are marked with a cross.}\label{fig:Weak-directed}
\end{figure}

\textbf{Weak directed edge} Figure \ref{fig:Weak-directed} shows the impact of a decreasing $\epsilon$ for constraining $X_1 \rightarrow X_4$. Notably, $\epsilon \leq 0.03$ does not almost achieve identification of $P(X_4=1|do(X_1=0,X_3=0)$ and $P(X_4=1|do(X_1=0,X_3=1)$, as the crosses indicate invalid upper bounds that were tightened below the true causal effect (dotted line). This demonstrates an $\epsilon$ choice assuming a too weak $X_1 \rightarrow X_4$.

\begin{figure}
  \centering
  \includegraphics[width=1\linewidth]{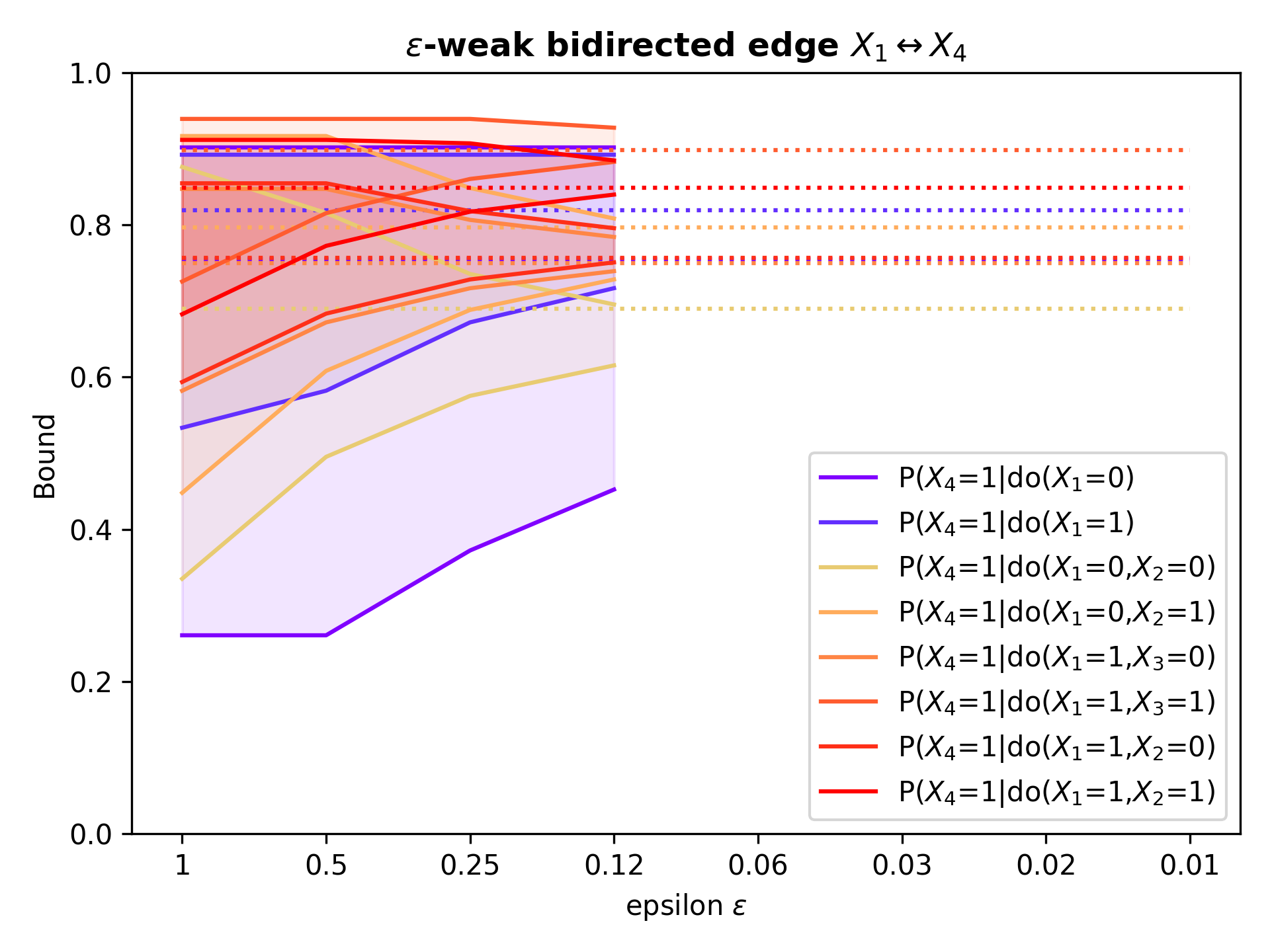}
  \caption{A weak bidirected edge tightens bounds. Lower epsilon values are falsified by infeasability.}\label{fig:Weak-bidirected}
\end{figure}

\textbf{Weak bidirected edge} Figure \ref{fig:Weak-bidirected} shows the impact of assuming an $\epsilon$-weak $X_1 \leftrightarrow X_4$. The linear program is infeasible from $\epsilon < 0.12$, i.e. using the given data the model falsifies the assumption of such a strong choice of $\epsilon$.

\begin{figure}
  \centering
  \includegraphics[width=1\linewidth]{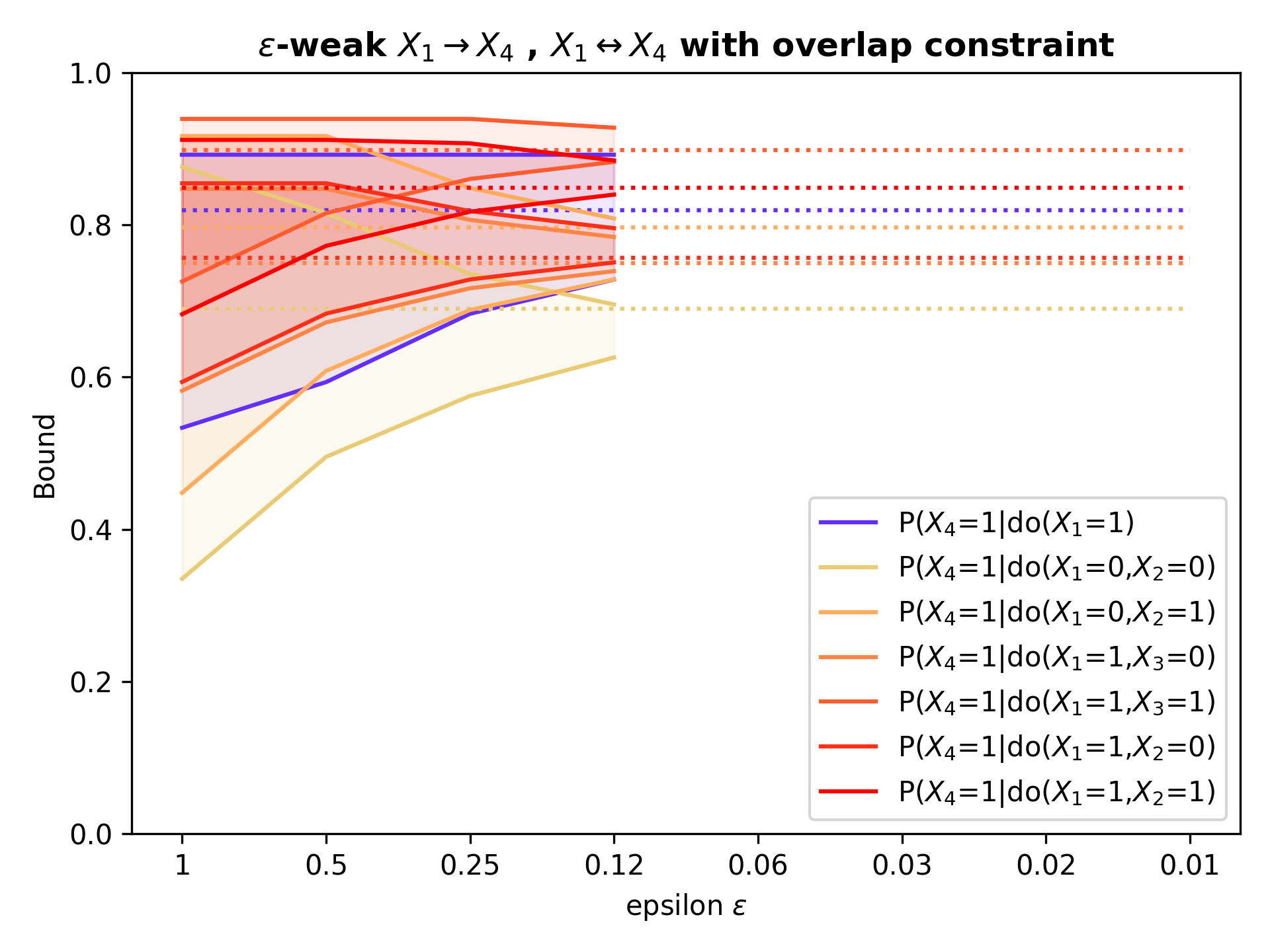}
  \caption{Weak directed and bidirected edges together with overlap constraint.}\label{fig:Weak-directed-bidirected-overlap}
\end{figure}

\begin{figure}
  \centering
  \includegraphics[width=1\linewidth]{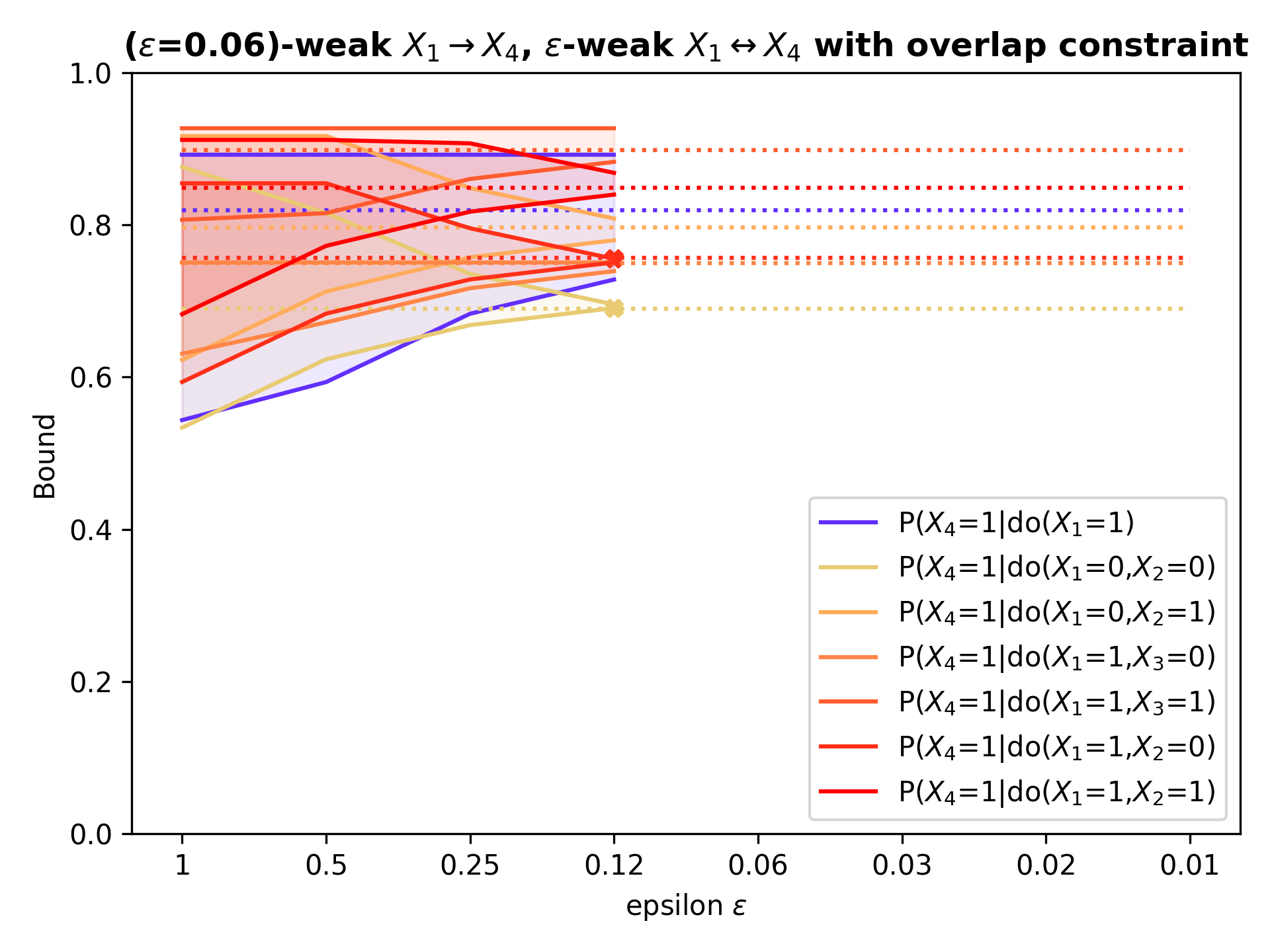}
  \caption{Bounds with all constraints in the model, with $\epsilon^{(X_1 \rightarrow X_4)} = 0.06$ and $\epsilon^{(X_1 \leftrightarrow X_4)}$ decreasing x-axis.}\label{fig:customepsilon}
\end{figure}

\textbf{Overlap with weak directed and bidirect edges} Figure \ref{fig:Weak-directed-bidirected-overlap} finally shows all three constraints together. The single intervention $P(X_4=1|do(X_1=1)$ has been slightly tightened further compared to Figure \ref{fig:Weak-bidirected} while $P(X_4=1|do(X_1=0)$ is not shown anymore, i.e. the combination of the constraints tightened it to a space in the polytope where choices of $\epsilon \geq 0.12$ do not affect it. For simplicity of exposition, we set $\epsilon$ to the same number for each weak edge constraint, but we can choose $\epsilon$ individually for each weak edge (e.g. $\epsilon^{(X_1 \rightarrow X_4)}$ and $\epsilon^{(X_1 \leftrightarrow X_4)}$). Figure \ref{fig:customepsilon} shows an example where we set $\epsilon^{(X_1 \rightarrow X_4)} = 0.06$ and vary $\epsilon^{(X_1 \leftrightarrow X_4)}$, leading to more tightening, but also two invalid bounds at $\epsilon = 0.12$.

\begin{figure}
  \centering
  \includegraphics[width=1\linewidth]{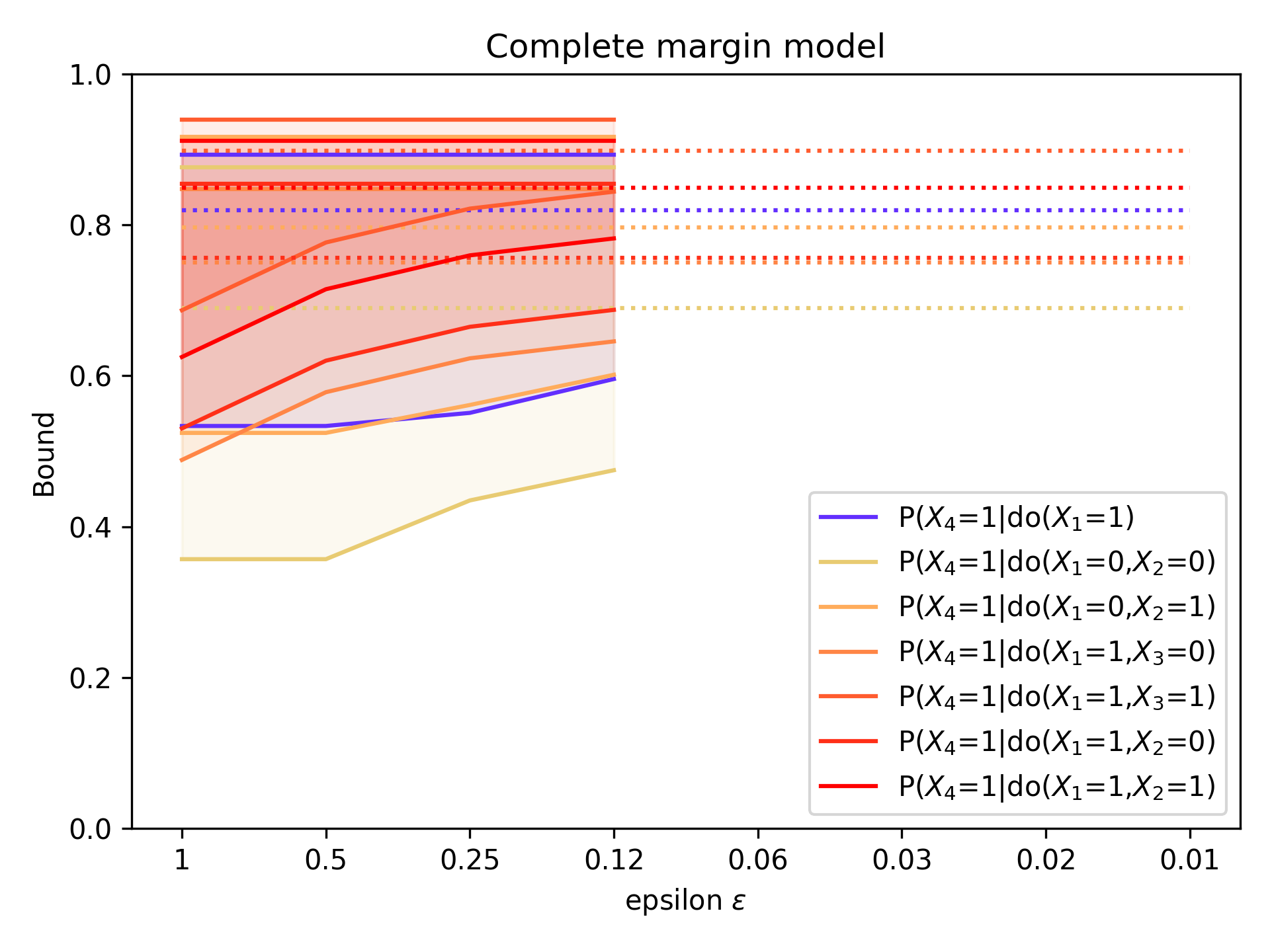}
  \caption{All constraints on the full margin of $\mathcal M_{global}$ in Figure \ref{fig:n4_DAG}. }\label{fig:n4_complete_margin_model}
\end{figure}

\textbf{Comparing with full parameterisation} Figure \ref{fig:n4_complete_margin_model} shows a causal marginal polytope with a single margin of size 4, i.e.  building a model directly from $\mathcal M_{global}$. With $2^{15}$ parameters it is generally the only full parameterisation of bounding model that is still computationally feasible. At a first glance, the bounds seem more loose compared to the causal marginal polytope. Importantly, such a comparison needs to be carefully done  as choices of $\epsilon$ have unique interpretations for each margin, including the full margin $\mathcal M_{global}$, as mentioned in the introduction to this Section. 

\subsection{Experiments with \texorpdfstring{$n>4$}{n>4}}
\label{sec:experimentswithnbigiger4}

\begin{figure}
  \centering
  \includegraphics[width=1\linewidth]{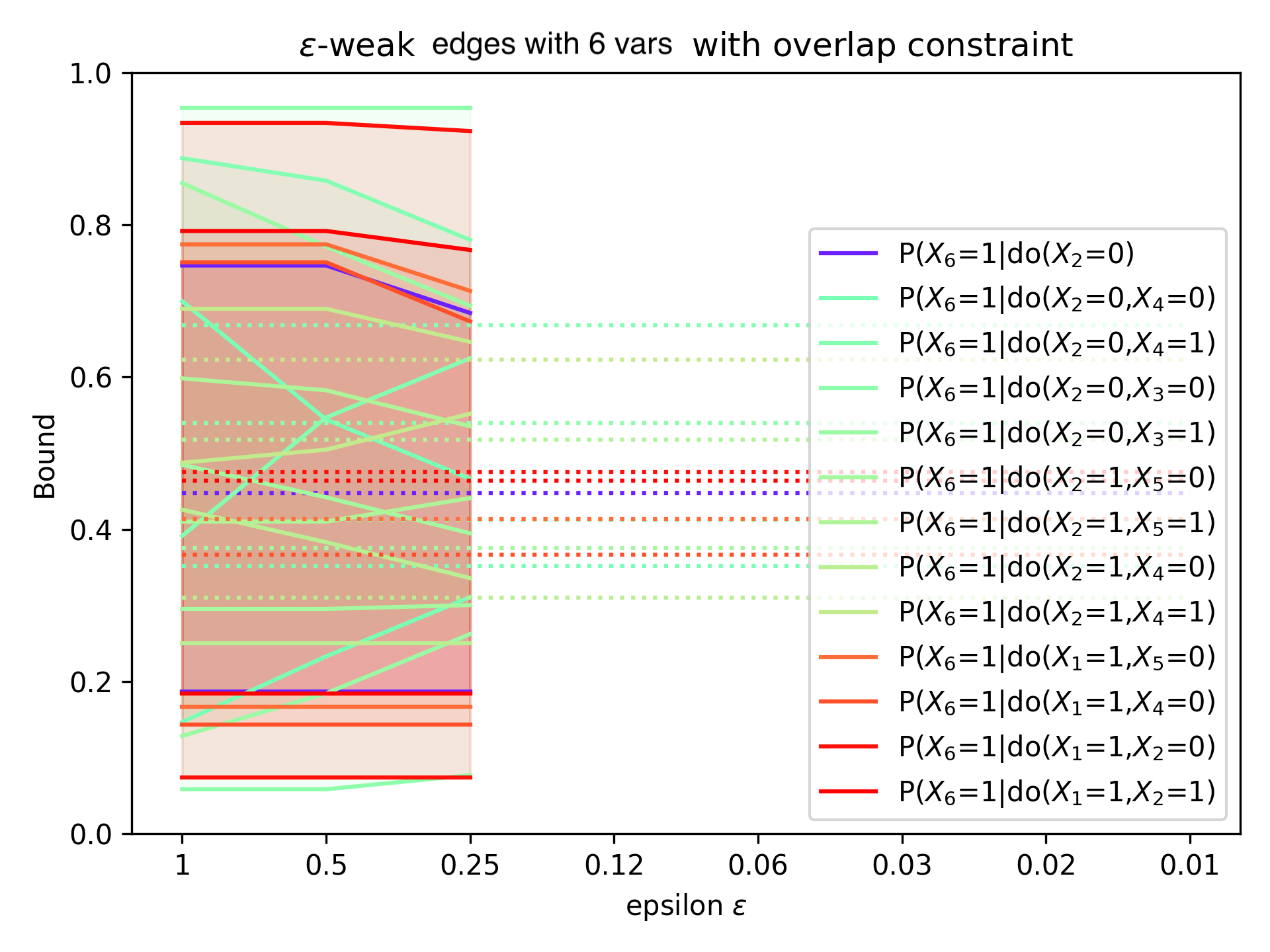}
  \caption{Bounds on 6 variable graph.}\label{fig:complete_n6}
\end{figure}

For a $n=6$ SCM , Figure \ref{fig:complete_n6} shows bounds on interventions assuming $\epsilon$-weak edges $X_1 \rightarrow X_4$, $X_1 \rightarrow X_6$, $X_3 \rightarrow X_5$, $X_1 \leftrightarrow X_4$ and $X_2 \leftrightarrow X_6$. We provide the same data regimes as in the previous example.
A comparison with a full margin model is computationally infeasible as it would require a model with $2^{31}$ parameters, compared to the computationally feasible $2^{1+2+4} \times 20 = 2560$ parameters of our causal marginal polytope with 20 margins each $|M_i| = 3$. In practice, an SCM with 10 variables will require only 15360 parameters and, given careful choice of the $\epsilon$-weak edges, can yield informative bounds.
Importantly, the examples above show causal systems where all three constraints are relevant to the margins, which is not always the case. Furthermore, the available data impacts the tightness of the bounds, with double intervention regimes being more useful for informing bounds on unobserved single intervention effects. Finally, to achieve the tightest bounds on the causal marginal polytope, a thorough evaluation should always individually choose $\epsilon$  for each constraining of weak directed and bidirected edges.

\section{Conclusion}
\label{sec:conclusion}

We introduced the causal marginal polytope to address two needs: computational scalability and an alternative way to elicit causal knowledge that will not require thinking in terms of a complete system. There are several uses which were not considered in this manuscript. For instance, it can guide experimental design where a small combination of design points can lead to information about combinations of promising smaller-dimensional (and hopefully cheaper) interventions. How to extend this idea to continuous variables, as a way of reducing variance in the complex Monte-Carlo based methods of e.g. \cite{kilbertus:2020,hu:2021,xia:2021}, is another direction of research.

\bibliography{rbas}

\end{document}